\documentclass[a4paper,twoside]{article}

\usepackage{ifpdf}
\ifpdf
	\usepackage[hidelinks]{hyperref}
\fi
\usepackage[T1]{fontenc} 
\usepackage{selinput}\SelectInputMappings{adieresis={ä},germandbls={ß}}
\usepackage[ngerman, english]{babel} 
\usepackage{scrhack} 
\usepackage{lmodern}
\usepackage[babel]{microtype} 
\usepackage{textcomp} 

\usepackage{xcolor} 

\usepackage[author=aprocha,color={0 0.87 1}]{pdfcomment} 

\usepackage{amsmath}
\usepackage{amssymb}
\usepackage{calc}
\usepackage{amstext} 
\usepackage{amsthm} 

\usepackage{algorithm}
\usepackage[noend]{algpseudocode}
\makeatletter
\def\BState{\State\hskip-\ALG@thistlm}
\makeatother
\makeatletter
\newcommand{\StatexIndent}[1][3]{%
	\setlength\@tempdima{\algorithmicindent}%
	\Statex\hskip\dimexpr#1\@tempdima\relax}
\makeatother

\usepackage{blindtext}

\usepackage{nicefrac} 
\usepackage{units} 

\usepackage[super]{nth}

\usepackage{pstricks}
\usepackage{graphicx}
\usepackage{import}
\usepackage{calc}
\usepackage[crop=pdfcrop]{pstool} 

\usepackage{tabularx} 
\usepackage{booktabs}
\usepackage{makecell} 

\usepackage[babel,german=quotes]{csquotes} 

\usepackage{epsfig}
\usepackage{caption}
\usepackage{subcaption}
\usepackage{multicol}
\usepackage{pslatex}
\usepackage{apalike}

\usepackage{cleveref}

\usepackage{SCITEPRESS}     

\newcommand{\hilodoe}{HiLoMoT-DoE}

\newcommand{\eg}[0]{e.~g.}
\newcommand{\ie}[0]{i.~e.}

\newcommand{\maxtext}{\text{max}}

\newcommand{\NRMSEF}{{\text{NRMSE}}}

\newcommand{\NRMSEValF}{\NRMSEF_{\text{val}}} 
\newcommand{\NRMSEVal}{$\NRMSEValF{}$}

\newcommand{\CVF}{{\text{CV}}_{K}} 
\newcommand{\CV}{$\CVF{}$}
\newcommand{\CVtenF}{{\text{CV}}_{10}} 
\newcommand{\CVten}{$\CVtenF{}$}
\newcommand{\CVhighF}{{\text{CV}}_{10\text{, high}}} 
\newcommand{\CVhigh}{$\CVhighF{}$}

\newcommand{\nMeasF}{{n}_{\text{Meas}}} 
\newcommand{\nMeas}{$\nMeasF{}$}

\newcommand{\SNRF}{\textrm{SNR}}
\newcommand{\SNR}{$\SNRF{}$}

\DeclareMathOperator{\yHat}{\mathnormal{\hat{y}}}
\DeclareMathOperator{\varHat}{{\hat{\sigma}}^{2}}
\DeclareMathOperator{\varHatM}{{\hat{\sigma}}^{2}_{\mathnormal{m}}}
\DeclareMathOperator{\varN}{{\sigma}^{2}_{\mathnormal{n}}}

\newcommand{\xHat}{\hat{x}}
\newcommand{\xAst}{x^{\ast}}
\newcommand{\xAstM}{x^{\ast}_{m}}
\newcommand{\ofX}{(\xHat)}
\newcommand{\ofXAst}{(\xAst)}
\newcommand{\ofXX}{(\xHat,\xHat)}
\DeclareMathOperator{\kHat}{\hat{\text{\textbf{k}}}}
\DeclareMathOperator{\covOp}{\mathnormal{k}}

\newcommand{\yHatOfX}{\yHat\ofX}
\newcommand{\varHatOfX}{\varHat\ofX}
\newcommand{\varMOfXAst}{\varHatM\ofXAst}
\newcommand{\covWithNoise}{(K+\varN I)}
\newcommand{\covOfXX}{\covOp\ofXX}
\newcommand{\yMatrix}{\text{\textbf{y}}}

\DeclareMathOperator*{\argmax}{arg\,max}

\newcommand{\isep}{\mathrel{{.}\,{.}}\nobreak}

\begin{document}
	\selectlanguage{english}

\title{Active Output Selection Strategies for Multiple Learning Regression Models}
\author{\authorname{Adrian~Prochaska\sup{1}\orcidAuthor{0000-0003-2707-1266}, Julien~Pillas\sup{1} and Bernard~Bäker\sup{2}}
\affiliation{\sup{1}Mercedes-Benz AG, 71059 Sindelfingen, Germany}
\affiliation{\sup{2}TU Dresden, Chair of Vehicle Mechatronics, 01062 Dresden, Germany}
\email{\{adrian.prochaska, julien.pillas\}@daimler.com, bernard.baeker@tu-dresden.de}
}


\abstract{
Active learning shows promise to decrease test bench time for model-based drivability calibration.
This paper presents a new strategy for \textit{active output selection}, which suits the needs of calibration tasks.
The strategy is actively learning multiple outputs in the same input space.
It chooses the output model with the highest cross-validation error as leading.
The presented method is applied to three different toy examples with noise in a real world range and to a benchmark dataset.
The results are analyzed and compared to other existing strategies.
In a best case scenario, the presented strategy is able to decrease the number of points by up to 30\,\% compared to a sequential space-filling design while outperforming other existing active learning strategies. 
The results are promising but also show that the algorithm has to be improved to increase robustness for noisy environments.
Further research will focus on improving the algorithm and applying it to a real-world example.
}

\onecolumn \maketitle \normalsize \setcounter{footnote}{0} \vfill

\section{Introduction}
\label{sec:intro}

Active learning methods -- sometimes called \textit{online design of experiments} or \textit{optimal experimental design} -- increase the capabilities of algorithms taking part in test design and execution \cite{cohn_neural_1996}. 
They reduce the required number of measurements significantly, while guaranteeing adequate model qualities \cite{klein_adaptive_2013}.
However, most methods aim at optimally identifying only one model. 
In most real-world applications, there are not one but multiple outputs. 
That leaves the test engineer with a question: 
Should all models be learned sequentially or simultaneously? 
And if they learn simultaneously, how to decide which model is the leading one?
Drivability calibration applications can be further distinguished from other active learning tasks because 
\begin{itemize}
\item the goal is to identify all measured outputs equally well and
\item pulling one query reveals the values of all outputs of interest.
\end{itemize}

\cite{dursun_ansatz_2015} showed a comparison of a sequential and a round-robin learning strategy for a drivability calibration task.
To the authors knowledge, no other publication analyses more sophisticated strategies for multiple learning regression models, which follow the conditions described above.
This paper proposes a new concept of learning strategy, which decides on the leading output by evaluating a cross validation error. 
This new strategy is compared to other existing strategies.
Multiple toy examples are used to create a noisy but reproducible test environment with different complexities.
At last, the strategy is also applied to a benchmark dataset. 

The paper is structured in six sections.
\Cref{sec:previousWork} of this paper introduces previous works in context of active learning in general and in particular for regression tasks. 
\Cref{sec:problemDefinition} focuses on describing the specialties of active learning in the calibration context. 
A new active learning task called \textit{active output selection} (AOS) is introduced there.
\Cref{sec:approach} describes the analyzed approaches. 
Furthermore, a new approach for AOS is presented. 
The approaches are evaluated using a toy example and a benchmark dataset.
Experimental details and a discussion of results are shown in \cref{sec:experiments}.
At the end, \cref{sec:conclusion} concludes the results and presents fields of possible future works.

\section{Previous work}
\label{sec:previousWork}

The field of active learning is a growing branch of the very present machine learning domain.
It is also referred to as \textit{optimal experimental design} \cite{cohn_neural_1996}.
\cite{settles_active_2009} shows a broad overview of the current state of the art in this discipline and gives an outlook to multiple possible future work fields.
Recent methodological advances in the scientific community mainly focused on classification problems. 
The main application domains are speech recognition and text information extraction \cite{settles_active_2009}.

While regression tasks in the context of active learning have not been as popular, the methodological development is relevant as well.
\cite{sugiyama_active_2008} propose an approach which actively learns multiple models for the same task and picks the best one to query new points.
\cite{cai_maximizing_2013} introduced an approach which uses expected model change maximization (EMCM) to improve the active learning progress for gradient boosted decision trees, which was later extended to choose a set of informative queries and to gaussian process regression models (GPs) by \cite{cai_batch_2017}.
\cite{park_robust_2020} propose a learning algorithm based on the EMCM, which handles outliers more robustly than before.
Those publications focus on new criteria for single output regression models to improve the active learning process. 
\cite{zhang_near-optimal_2016} present a learning algorithm for multiple-output gaussian processes (MOGP) which outperforms multiple single-output gaussian processes (SOGP). 
However, this publication focuses on improving the prediction accuracy of one target output with the help of several correlated auxiliary outputs. The experiments indicate that a global consideration is beneficial.

There were also advances in active learning for automotive calibration tasks for which the identification of multiple process outputs in the same experiment is more relevant to the application.
\cite{klein_adaptive_2013} applied a design of experiments for hierarchical local model trees (\hilodoe{}), which was presented by \cite{hartmann_adaptive_2013}, successfully to an engine calibration task.
They presented two application examples with two outputs each and five respectively seven inputs. 
The two outputs were modeled with a sequential strategy, which identifies an output model completely before moving to the next one \cite{klein_adaptive_2013}.

\cite{dursun_ansatz_2015} applied the \hilodoe{} active learning algorithm to a drivability calibration example characterized by multiple static regression tasks with identical input spaces.
They analyzed the sequential strategy already shown by \cite{klein_adaptive_2013} and compared it to a round-robin strategy, which switches the leading model after each iteration/measurement \cite{dursun_ansatz_2015}.
The authors show that the round-robin strategy outperforms offline methods and the online sequential strategy in this experiment. 
It might indicate, that round-robin is preferably used in general, but further experiments are necessary.
Since then, no efforts have been made to analyze active learning strategies for multiple outputs.

\section{Problem definition}
\label{sec:problemDefinition}

The analyses of this paper are motivated by the field of model-based drivability calibration.
For this application, an active learning algorithm learns a number of $M$ different outputs, which are possibly non-correlated.
Their models are equally important for succeeding optimizations, so the goal is to identify adequate models for all outputs.
The input dimensions of all models are the same.
Querying a new instance corresponds to conducting a measurement on powertrain test benches.
Therefore, a measurement point is cost-sensitive, which is inherent to active learning problems.
Contrary to other applications, every single measurement provides values for all $M$ outputs\footnote{This is in contrast to \eg{} geostatistics, where measuring any individual output, even at the same place (\ie{} model inputs), has its own costs \cite{zhang_near-optimal_2016}.}.
Tasks of simultaneously learning $M>1$ process outputs with equal priority and multi-output measurements are not known in the scientific community.
In the following, they are referred to as \textit{active output selection} (AOS).
All measured outputs contain to some extent noise.
The signal-to-noise-ratio $\SNRF\!_{m}$ of model $m$ is the ratio between the range of all measurements $y_{m}$ and the standard deviation $\sigma_{\textrm{N}}$ of normally distributed noise: $\SNRF\!_{m}=\frac{\max\left({y_{m}}\right)-\min\left({y_{m}}\right)}{\sigma_{\textrm{N}}}$. 
The \SNR{} for drivability criteria lies approximately in a range of $\left(7\isep{} 100\right)$ and can be different for each criterion.

For applications on a test bench, conducting a measurement is timely more expensive than the evaluation of code.
This is why the performance of code is not crucial in this context and is only discussed openly in this paper instead of analyzing it systematically.

\section{Active output selection strategies}
\label{sec:approach}

This paper analyzes strategies for AOS tasks with $M>1$ regression models.
In this paper, each of those $M$ process outputs is modeled with a GP since they handle noise in the range of vehicle calibration tasks very robustly \cite{tietze_model-based_2015}.
The leading process output defines the placement of the query in each iteration.
A simple maximum variance strategy is deployed as active learning algorithm: 
A new query $\xAstM$ in the input space $\mathbb{X}$ is placed at that point, where the output variance is maximal.
\begin{equation}
\xAstM = \argmax_{\xAstM \in \mathbb{X}}{\left( \varMOfXAst \right) } \label{eq:maxVarianaceStrategy}
\end{equation}
This approach was presented by \cite{mackay_information-based_1992} for general active learning purposes and applied and evaluated on GPs \eg{} by \cite{brauer_gaussian_2000} or \cite{pasolli_gaussian_2011}. 
The implementation of such a learning strategy is straightforward for GPs since the output variance at each input point is directly calculated in the model. 
\Cref{eq:GPoutput} and \cref{eq:GPvariance} show the calculations of the predicted mean $\yHat$ and output variance $\varHat$ of a GP. 
$\kHat$ is the vector of covariances $\covOp(X,\xHat)$ between the measured training points $X$ and a single test point $\xHat$, $K=K(X,X)$ are the covariances of $X$ and $\yMatrix$ contains the observations under noise with variance $\varN$ \cite{rasmussen_gaussian_2008}.
\begin{align}
\label{eq:GPoutput}
\yHatOfX &= \kHat^{T} \, \covWithNoise^{-1} \, \yMatrix
\\
\label{eq:GPvariance}
\varHatOfX &= \covOfXX - \kHat^{T} \covWithNoise^{-1} \kHat
\end{align}

Depending on the AOS strategy the leadership of the learning process is chosen differently. 
In the following, three already existing and one new active learning strategy (CVH) as well as a passive sequential design are described. All of those strategies are empirically analyzed in \cref{sec:experiments}.

\paragraph{sequential strategy (SQ)}
After measuring a set of initial points, the first process output is leading. 
When the desired model accuracy or the maximum number of points is reached, the next model places measurements and is identified. 
This procedure is repeated until the criteria for all $M$ models are fulfilled.
The maximum number of measurements for every $i$-th model is calculated as follows:
\begin{equation}p_{m,\maxtext{}}=\frac{p_{\maxtext{}}-p_{\text{init}}}{M}\end{equation}

An advantage of SQ is, that it identifies only one model each iteration.
Depending on the complexity and noise of all process outputs, the order of leading models might influence the performance of this strategy.

\paragraph{round-robin strategy (RR)}
This strategy changes the leading model after each measurement. 
Models that have reached the desired model quality are not leading any longer.
An advantage of round-robin is, that the order of process outputs only has a very small influence on planning the measurements, since the models are switched with every step. 
Therefore, this strategy should be more suited to handle tasks where the outputs have different complexities. 
RR also identifies only one model each iteration.

\paragraph{global strategy (G)}
This strategy chooses that query $x^{\ast}$, which maximizes the sum of output variances.
\begin{equation}
\xAst = \argmax_{\xAst\in \mathbb{X}} {\left( \sum_{m=1}^{M} {w_{m} \varMOfXAst} \right) }
\end{equation}
This is a weighted compromise between all models with the weights being $w_m=1$.
G identifies all $M$ models each iteration and is therefore computationally more expensive than SQ and RR.

\paragraph{\CVhigh{} strategy (CVH)}
\Cref{alg:CVH} shows the \CVhigh{} strategy.
In the beginning, CVH plans the queries of an initial set of points and conducts the measurements.
Afterwards, CVH identifies the models of all outputs in each iteration.
Additionally, the model errors are calculated.
In this case, a model error is expressed using the normalized root mean squared $K$-fold cross-validation-error \CV{} with $K=10$. 
\Cref{eq:CVError} shows the general form of \CV{} of the $m$-th model with the predictions ${\hat{y}^{-\kappa(i)}}_{m,i}$ of the $m$-th model being identified without measurements of set $\kappa:\left\{1,\dots,N\right\} \mapsto \left\{1,\dots,K\right\}$
\begin{align}
\label{eq:CVError}
\CVF{}\!_{,m} = \sqrt{\frac{\sum_{i=1}^{N}{\left ( {y}_{m,i} - {\hat{y}^{-\kappa(i)}}_{m,i} \right)^{2}} }{\max\left({y}_{m}\right)-\min\left({y}_{m}\right)}}
\end{align}
The usage of another accuracy or error criterion is possible, but \CV{} is well-comparable between models.
For stability reasons, \CVten{} is filtered with a digital moving average filter, which reduces the influence of fluctuations during runtime. 
In every following iteration, the output with the highest model error is leading the learning process.
This output is assumed to benefit the most from being in leadership of learning. 

\begin{algorithm}[h]
	\caption{CVH active output selection strategy.}\label{alg:CVH}
	\begin{algorithmic}[1]
		\Repeat
		\If {no initial points have been carried out}
		\State plan queries of initial points
		\Else
		\State find model with the highest filtered 
		\StatexIndent[2] cross-validation error
		\State calculate next query
		\EndIf
		\State conduct measurements on planned queries
		\ForAll{models}
		\State update model
		\State assess cross-validation error
		\State filter the cross-validation error
		\EndFor
		\Until{maximum number of points or desired model quality is reached}
	\end{algorithmic}
\end{algorithm}

Using \CVten{} obliges identifying each of the $M$ models for $10$ times in each iteration. 
Compared to the other strategies, this results in a higher computational effort than the previously presented methods.
However this argument is not crucial for drivability calibration tasks, as the measurements itself take a lot longer than calculating the succeeding query.
Since \CVten{} also increases with higher noise, this strategy might be prone to one process output with significantly larger noise than the others.
Its model cannot reach a model error as low as those of the other outputs; after reaching the minimum possible \CVten{} the model will not benefit from actively planning points anymore.
To the authors' knowledge, this strategy has not been presented or analyzed in any other publication.

\paragraph{sequential space-filling strategy (passive, SF)}
Instead of a random sequential set of queries, the authors choose including a passive but sequential design as baseline method to verify the benefits of those AOS strategies.
This kind of design is derived from an s-optimal (space-filling)  experimental design, which is preferred over a random set of points in drivability calibration applications.
A sequential method additionally enables a fair comparison on whether an active learning strategy is truly beneficial over a passive one.
After an initial set of measurements, the next point is always placed in a maximin-way which maximizes the minimum Mahalanobis-distances $d_{\min}$\footnote{The Mahalanobis distance for uncorrelated data in a range between 0 and 1 is identical to the Euclidian distance.} between a huge set of candidate points and the already measured points.
\begin{align}
d_{\min}{(\xHat)}=\min{\left\|{\xHat}-{X}\right\|}
\\
x^{\ast} = \argmax_{\xAst\in \mathbb{X}}\left(d_{\min}(\xHat)\right)
\end{align}

Because points are planned sequentially, this design does not exactly result in a test design which is optimally space-filling for the current number of points.
However, it is an easy way to be close to this optimality during a sequential design where the number of points is not predefined.

\paragraph{}
Due to the characteristics of the AOS strategies described above, the following hypothesis are tested with the experiments:
\begin{enumerate}
	\item The non-heuristic CVH is in many cases beneficial but also has drawbacks concerning high noise-induced generalization error. 
	\item RR is robust in all use cases but can be outperformed by CVH.
	\item The active learning strategies perform significantly better than a SF.
\end{enumerate}

\section{Experiments}
\label{sec:experiments}
The application of the presented learning strategies in the field of drivability calibration is designed for the use on a test bench.
However, typical static drivability criteria have a signal-to-noise-ratio \SNR{} of $\left(7\isep{} 100\right)$.

\begin{figure}[!h]
	\centering
	\includegraphics{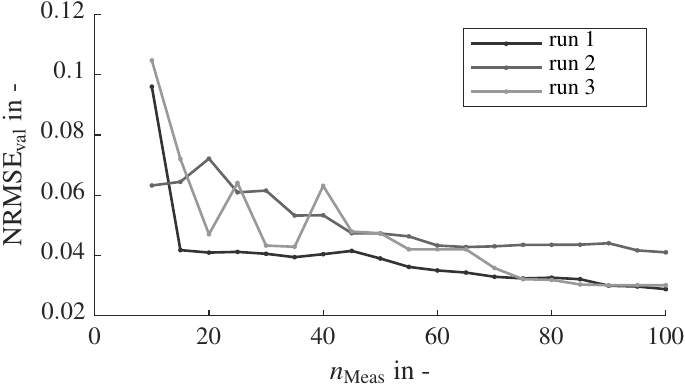}
	\caption{${\text{NRMSE}}_{\text{val}}$ for 3 runs of a generic process with a space-filling design and different noise observations. This figure shows the influence of noise on \NRMSEVal{} for a space-filling strategy.}
	\label{fig:noiseInfluence}	
\end{figure}

\Cref{fig:noiseInfluence} demonstrates the influence of noise in that range in a toy example.
It shows the \NRMSEVal{}-values over the number of measurements \nMeas{} of 3 learning procedures of a space-filling design for the same example.
The normalized root mean squared error of the validation points \NRMSEVal{} is calculated according to \cref{eq:NRMSEVal}.
\begin{equation}
	\label{eq:NRMSEVal}
	\NRMSEValF\!_{,m}=\sqrt{\frac{\sum_{i=1}^{N_{\text{val}}}{\left ( {y}_{m,i\text{,val} } - {\hat{y}}_{m,i\text{,val}} \right)^{2}} }{\max\left({y_{m}}\right)-\min\left({y_{m}}\right)}}
\end{equation}
The only difference between those runs are the locations of initial points and the noise observations, which are characterized by $\SNRF{}=12$. 
This amount of noise leads to completely different generalization errors.
Those results propose that to compare learning strategies for noisy environments, an experiment has to be repeated multiple times.
For comparing the results of different learning strategies, not only the mean but also the standard deviation of \NRMSEVal{} is relevant.
The conduction of numerous tests for example on a powertrain or engine test bench is time- and cost-expensive and therefore not practicable.
Furthermore, the test conditions would be slightly different every time which makes a direct comparison difficult.
The authors’ goal in this publication is to compare the described learning strategies in a way that is reproducible and representative for applications in vehicle calibration. 
That is why three different toy examples are chosen for comparison here.
They use the possibility of computer generated noise to be reset to the same starting point of the random-number-generator.
\subsection{Toy Examples}
\label{sub:toyExamples}

Every toy example includes three different generic processes, which are analytical multi-dimensional sigmoid or polynomial models generated by a random function generator presented in \cite{belz_proposal_2015}.
Their outputs are overlaid with normally distributed noise to simulate the measurement inaccuracy.
Each process has two input dimensions.
The setup used for comparisons consists of multiple runs.
One run is understood as one single observation for those comparisons.
The overlaid noise of one run is the same for each learning strategy.
This is a condition that real world tests cannot fulfill – or at least with untenable effort.
However, it increases comparability: Each strategy has the same initial conditions for each run.
Furthermore inside one run, the chronological order of examined process outputs and the randomly created, initial points are the same for each strategy.

Another advantage of a comparison with analytical models is the knowledge about the real values from the underlying process without influences of noise.
All identified models are validated with 121 gridded validation points and the \NRMSEVal{} is calculated (see \cref{eq:NRMSEVal}).

The hypotheses stated earlier are analyzed with 3 different toy examples. 
For every toy example three different generic processes are chosen. 
This is a realistic experience value for the number of process outputs to be modeled.
The characteristics of those different setups are shown in \cref{tab:specification}. 
Every learning strategy is tested for 50 times in each setup. 

\begin{table}[h]
	\centering
\caption{Specification of the three setups. The symbol + is indicating high complexity or noise.}
\label{tab:specification}
	\begin{tabular}{@{\hskip3pt}crcc@{\hskip3pt}}
		\toprule
		setup & \begin{tabular}{@{}r@{}}
			model type of each \\ toy example output\\
			\end{tabular} 
		& complexity & noise \\
		\midrule
		 & sigmoid & + & +  \\
		1 & sigmoid & + & + \\
		& sigmoid & + & + \\ \addlinespace
		 & polynomial & $\circ$ & +  \\
		2 & polynomial & $\circ$ & +  \\
		& \begin{tabular}{@{}r@{}}
			multiple sigmoids\\
			combined with steps\\
			\end{tabular} 
		& ++ & + \\ \addlinespace
		 & sigmoid & + & ++ \\
		3 & sigmoid & + & + \\
		& \begin{tabular}{@{}r@{}}
	multiple sigmoids\\
	combined with steps\\
\end{tabular} 
& ++ & + \\
		\bottomrule
	\end{tabular}

\end{table}

For better understandability of the results the squared sum of  the \NRMSEVal{} of each model is used for comparison.
\begin{equation}
\label{eq:NRMSEValSum}
\NRMSEF_{\text{val,}\varSigma} = \sqrt{\sum_{m=1}^{M}{\left( \NRMSEF_{\text{val,}m}\right) ^{2}}} 
\end{equation}

\begin{figure*}[!h]
	\centering
	\includegraphics{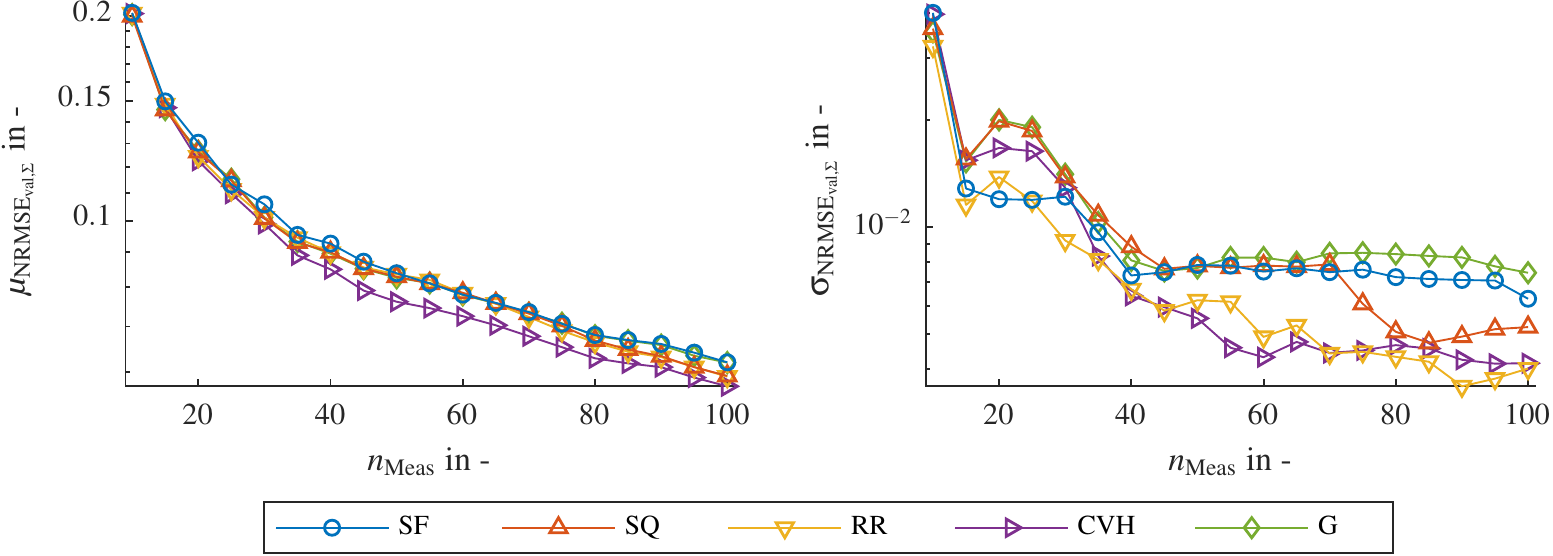}
	\caption{Mean $\mu$ and standard deviation $\sigma$ of the $\NRMSEF_{\text{val,}\varSigma}$ of setup 1 over the number of measurements \nMeas{}.}
	\label{fig:similarComplexity}	
\end{figure*}

\Cref{fig:similarComplexity} shows the results of setup 1. 
When $\nMeasF\lessapprox30$, the performance of the analyzed strategies are all very similar. 
From that point on, the new CVH strategy performs significantly better than all other strategies. 
This includes a low standard deviation $\sigma_{\NRMSEF_{\text{val,}\varSigma}}$.
The low $\sigma_{\NRMSEF_{\text{val,}\varSigma}}$ shows that there is not a big difference between different runs and therefore stands for the high robustness of this method.

The mean of CVH has the lowest end value. 
This difference is significant compared to the results of RR and CVH, which have very similar means.
Assuming that we want to quit our experiment at any time, CVH should be preferred. 
RR and SQ have similar mean performance, but RR inherits a lower standard deviation and is therefore more robust.
Compared to the \NRMSEVal{}$\!_{,\varSigma}$ of SF after $\nMeasF=100$, RR and SQ reduce the number of points by \unit[5]{\%} whereas CVH reduces the number of points even further by \unit[15]{\%}.
The mean performance of G is comparable to SF.
However, the variance of G is higher, which ranks the performance of SF over G.

\begin{figure*}[!h]
	\centering
	\includegraphics{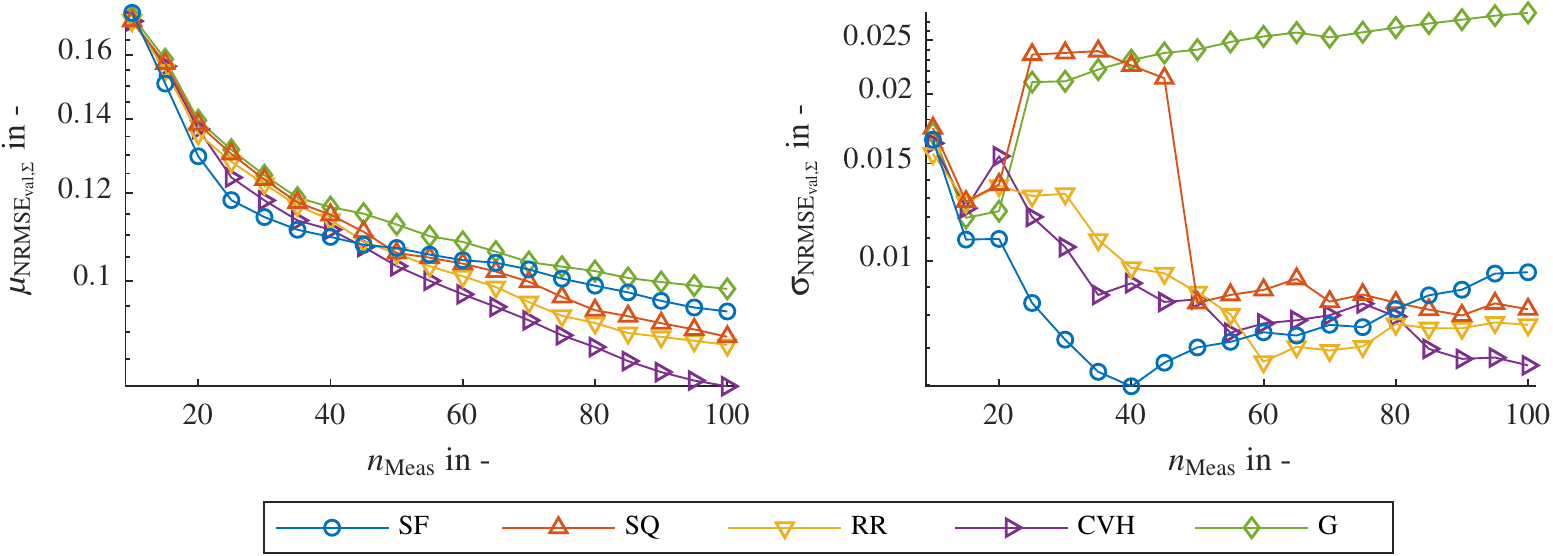}
	\caption{Mean $\mu$ and standard deviation $\sigma$ of the $\NRMSEF_{\text{val,}\varSigma}$ of setup 2 over the number of measurements \nMeas{}.}
	\label{fig:differentComplexity}	
\end{figure*}

\Cref{fig:differentComplexity} shows that the difference between active learning strategies is especially high in setup 2 compared to the SF design. 
In contrast to the previous setup, one process output has a much higher complexity than the other ones. 
This combination shows the benefits of CVH very clearly.
When $\nMeasF\gtrapprox45$, CVH performs better than all other strategies. 
G performs significantly worse than the other strategies. 
This is unlike the third hypothesis in \cref{sec:approach}.
Depending on the application and the chosen strategy, it is not always beneficial to use active learning.
After $\nMeasF=100$, RR and SQ have no significant difference in results. 
However, the standard deviation of SQ is higher during the runs, especially for $\nMeasF\lessapprox45$. 
The authors assume that the influence of the process output order plays a role in that. 
Compared to that, RR shows a more robust behavior.
RR reaches the end value of SF at $\nMeasF=75$.
The standard deviations of RR and CVH are both on similar levels.
The CVH performs significantly better than all other strategies in this scenario.
It reduces the number of measurement points of a SF by \unit[30]{\%} and reaches the end value of RR after $\nMeasF=80$.

\begin{figure*}[!h]
	\centering
	\includegraphics{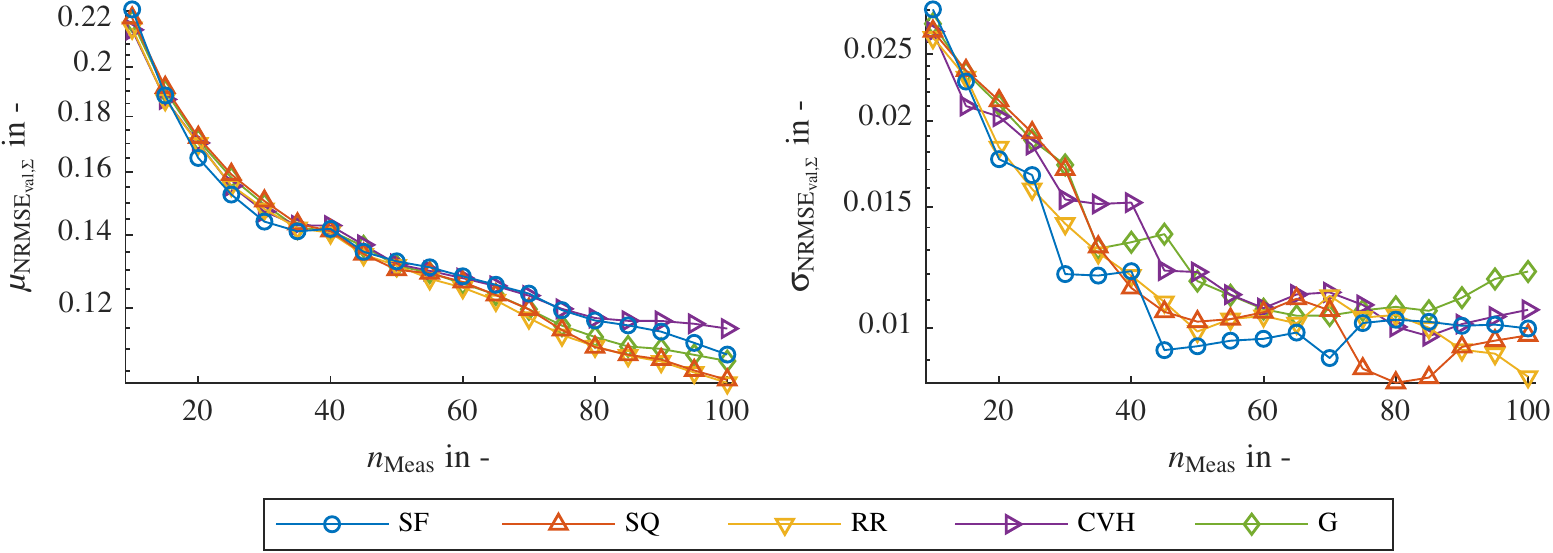}
	\caption{Mean $\mu$ and standard deviation $\sigma$ of the $\NRMSEF_{\text{val,}\varSigma}$ of setup 3 over the number of measurements \nMeas{}.}
	\label{fig:differentNoise}	
\end{figure*}

\Cref{fig:differentNoise} shows the results of setup 3.
Those results match the expectations that CVH performs worse in an environment where a process output with high noise and a complex one exist.
For $\nMeasF{}\lessapprox80$, the overall performance of CVH is not much worse than other strategies.
However, all other strategies have a better end value.
In this setup, where there is one very noisy and one very complex process output, RR and SQ perform best.
They outperform SF at $\nMeasF=85$ and CVH already at $\nMeasF=75$.
Compared to SQ, RR is rather robust in the beginning and in the end.

The experiments confirm the hypotheses stated in \cref{sec:approach}.
Only the third hypothesis turned out not to be true in all cases: SF can outperform active learning in some cases.

\subsection{Benchmark dataset}
\label{sub:benchmarkDataSet}

As stated in \cref{sec:problemDefinition}, applications of active learning in the domain of drivability calibration are rather unique concerning the conditions and goals of other existing tasks.
That is, why there is no benchmark dataset that fully suits the needs of an example.
However we wanted to demonstrate the practical use of such a learning strategy.
This is why the jura dataset \cite{goovaerts_geostatistics_1997}, which is actually a dataset from the domain of geostatistics, is used as a benchmark dataset here.
This dataset contains the concentration of 7 heavy metal concentrations at different locations in the Swiss Jura.
It is the best fitting dataset, which is also used to evaluate the learning algorithms in \cite{zhang_near-optimal_2016}.
In contrast to that publication, we set the goal to model every of the three chosen outputs equally well.
Three concentrations (Ni, Cd, Zn) are modeled as a function of the locations during every test run.
The results of the AOS strategies presented in \cref{sec:approach} are averaged over 50 test runs.

\begin{figure*}[!h]
	\centering
	\includegraphics{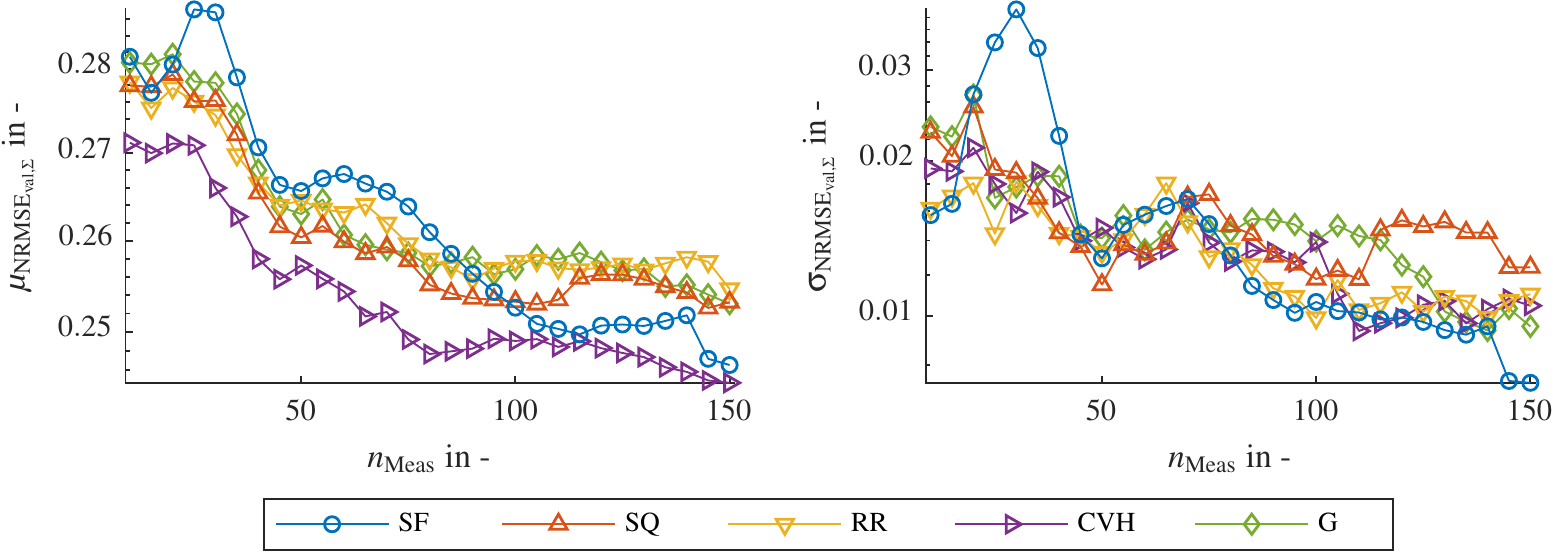}
	\caption{Mean $\mu$ and standard deviation $\sigma$ of the $\NRMSEF_{\text{val,}\varSigma}$ of the Jura Dataset over the number of measurements \nMeas{}.}
	\label{fig:juraDataset}	
\end{figure*}

\Cref{fig:juraDataset} shows the results of the benchmark dataset.
In the beginning, SF performs significantly worse than CVH. 
After $\nMeasF\gtrapprox 100$, there are no significant differences between those two strategies.
CVH and SF outperform G, SQ and RR in the end.
Throughout all measurements however, the mean of CVH is the lowest.
Since experiments on a test bench might be stopped after any fixed number of measurements, the results indicate that CVH is preferably used, although it is not significantly better than SF regarding the final model accuracy.

\section{Conclusion}
\label{sec:conclusion}

In this paper, \textit{active output selection}, a new task for active learning is introduced.
It is characterized by identifying models of multiple process outputs with the same input dimensions.
We present a new strategy (CVH) to define the leading models in an active output selection setup. 
The decision is based on the cross-validation errors of the identified models.
The paper thoroughly analyzes the advantages and disadvantages of said strategy.
The process outputs are identified using gaussian processes (GP).
A simple maximum variance algorithm is chosen as active learning strategy for each individual output.
The strategy is analyzed on different toy examples, which include noisy generic process outputs.
The results of CVH are compared against three existing learning strategies: round-robin, sequential and global.
Furthermore a passive, sequential space-filling strategy is chosen as baseline for the active learning strategies.

The results show that the presented strategy is preferably used in most real-world setups.
The performance and robustness are good compared to other multi-output strategies.
Compared to the baseline strategy, CVH saves up to \unit[30]{\%} of the measurements. 
In this setup, which has one process output with higher complexity, CVH outperforms any other active output selection strategy.
In the particular case of a setup with one output with high noise and one output with high complexity, other strategies perform better than CVH.
The consideration of the estimated generalization error could further improve the performance of CVH especially for those setups and will therefore be context to further investigation.

The results of a benchmark dataset confirm the good performance of CVH in the toy examples.
Due to the lack of a more suitable public benchmark dataset however, a geostatistics example was chosen. 
A public benchmark dataset from the field of drivablity calibration would facilitate comparisons and simplify further work on the subject.
Another future research will apply the presented strategy to setups with different numbers of input dimensions.
Moreover the application of those strategies on different modeling and single model active learning approaches is promising.

\bibliographystyle{apalike}
{
	\bibliography{VO3-learner_choice}}

\end{document}